\documentclass[11pt,letterpaper]{article}
\usepackage[letterpaper,textwidth=6.75in,textheight=9in,top=1in]{geometry}
\usepackage{mathptmx}
\usepackage[round,authoryear]{natbib}

\usepackage[utf8]{inputenc}
\usepackage[T1]{fontenc}
\usepackage{url}
\usepackage{booktabs}
\usepackage{amsfonts}
\usepackage{amsmath}
\usepackage{amssymb}
\usepackage{nicefrac}
\usepackage{microtype}
\usepackage{xcolor}
\usepackage{graphicx}
\usepackage{subcaption}
\usepackage[ruled,vlined,linesnumbered]{algorithm2e}
\usepackage{placeins}
\usepackage{hyperref}
\hypersetup{colorlinks=true,linkcolor=blue!45!black,citecolor=blue!45!black,urlcolor=blue!45!black,
  pdftitle={RBS-Attention: Radius-Bounded Sparse Prefill for Long-Context Large Language Models},
  pdfauthor={Chuxu Song}}
\newcommand{\method}{RBS-Attention}

\title{\method: Radius-Bounded Sparse Prefill for Long-Context Large Language Models}

\author{
Chuxu Song$^{1}$ \quad
Jiuqi Wei$^{2}$ \quad
Zhencan Peng$^{1}$\thanks{Corresponding author.}\\[4pt]
$^{1}$Department of Computer Science, Rutgers University\\
$^{2}$OceanBase, Ant Group\\[4pt]
\texttt{\{chuxu.song, zhencan.peng\}@rutgers.edu}\\
\texttt{weijiuqi.wjq@antgroup.com}
}
\date{}

\begin{document}

\maketitle

\begin{abstract}
Long-context large language model inference is increasingly limited by prefill, where dense self-attention processes the entire prompt before generation begins. Sparse block selection can reduce this cost, but a block centroid may hide a highly relevant token among many irrelevant ones. We call this failure mode \emph{mean dilution} and propose \method{}, a training-free sparse-prefill method with two complementary selection branches. A centroid base branch captures average relevance, while a rescue branch uses the maximum key-block radius and its prompt-, layer-, and head-dependent distribution to identify blocks at risk of underestimation. Independently thresholding the two branches and combining their masks controls the contribution of rescue blocks while preserving regular block-sparse FlashAttention execution. On H100 GPUs, \method{} achieves 20.65$\times$ standalone prefill-attention speedup, 11.92$\times$ vLLM prefill-attention speedup, and 5.97$\times$ end-to-end time-to-first-token speedup at 128K on Qwen3-30B-A3B-Instruct-2507-FP8. On the dense Qwen3-32B model, it obtains 88.65 overall RULER accuracy versus 89.52 for dense attention; LongBench-v2, InfiniteBench, and Video-MME provide additional quality evaluation. Supporting experiments measure actual retention, compare selectors at matched density, and characterize block-size, threshold, and memory behavior. Together, these results support radius-adaptive dual-branch selection as an effective approach to long-context prefill.
\end{abstract}

\section{Introduction}
\label{sec:introduction}

Large language models (LLMs) increasingly process long contexts for repository-level code understanding~\citep{liu2024repobench,jimenez2024swebench}, multi-document question answering~\citep{yang2018hotpotqa,liu2024lostmiddle}, retrieval-augmented generation~\citep{lewis2020retrieval,izacard2021leveraging}, and multimodal reasoning~\citep{yue2024mmmu,fu2024videomme}. These applications may ingest tens or hundreds of thousands of tokens before generating an answer. The resulting \emph{prefill} stage is latency-critical: optimized dense-attention kernels improve memory access and hardware utilization~\citep{dao2022flashattention,dao2023flashattention2}, but still perform a quadratic number of query--key interactions.

Sparse-prefill systems reduce this work by inferring attention patterns or selecting coarse key-value blocks before computing token-level attention~\citep{jiang2024minference,fan2026flashprefill,lai2025flexprefill,xu2025xattention}. Their effectiveness depends on selecting useful evidence with substantially less work than dense attention. This is difficult when the evidence consists of a few tokens embedded in a heterogeneous block: a cheap aggregate can assign that block a low score even when an individual token is highly relevant.

We study this failure mode, which we call \emph{mean dilution}. A key-block centroid summarizes average alignment with a query. If one relevant key is surrounded by unrelated keys, its contribution to the mean is diluted and the selector may discard the entire block. Once pruned, the block cannot contribute to the sparse attention output. Our empirical analysis connects this underestimation to the maximum distance of a key from its block centroid. Centroid rank underestimation increases with block radius, and the highest-radius quintile contains 39.5\% of the top-5\% attention blocks in our diagnostic measurement (Figure~\ref{fig:observations}). This motivates using radius to identify blocks that warrant additional selection capacity.

We propose \method{}, a training-free selector that combines \emph{centroid relevance} and \emph{radius-adaptive rescue}. The base branch scores the centroid, retaining blocks with strong average relevance. The rescue branch adds a radius contribution scaled by a coefficient computed from the current prompt's radius distribution, separately for each layer and KV head. Compact blocks receive little rescue weight; highly dispersed blocks receive more. Each branch applies its own relative threshold, and their masks are combined with the forced sink, local-window, and recent blocks. The two thresholds control the amount of selection devoted to relevance and rescue, while the realized block density remains content-dependent.

The geometry underlying rescue is a centroid-centered L2 bound. Its role is to motivate a risk signal: the actual radius contribution is attenuated, so it need not remain a strict upper bound. This distinction also clarifies the relation to query-aware page retrieval such as Quest~\citep{tang2024quest}, which ranks decode-time pages using coordinate-wise min/max bounds. RBS uses a continuous, distribution-adaptive rescue weight together with a separately selected base branch to construct a two-dimensional prefill mask. Section~\ref{sec:related_work} discusses related approaches to query-aware and heterogeneous-block selection, and our supporting experiments compare the complete selector with centroid-only, Full-L2, and Quest-style alternatives.

After selection, RBS executes attention on the retained block pairs through a block-sparse FlashAttention kernel. The contribution lies in the radius-adaptive selection and the coordination of its two branches; relative-threshold selection and the sparse kernel build on existing prefill systems. This design preserves regular GPU execution while reducing the risk that centroid scoring removes a localized piece of evidence.

Our primary evaluation covers system performance on H100 GPUs and quality on dense, MoE, and multimodal Qwen models~\citep{yang2025qwen3,qwen2025qwen3vl}. On Qwen3-30B-A3B-Instruct-2507-FP8 at 128K context length, RBS achieves 20.65$\times$ standalone prefill-attention speedup, 11.92$\times$ vLLM prefill-attention speedup, and 5.97$\times$ end-to-end time-to-first-token (TTFT) speedup. On Qwen3-32B, it obtains 88.65 overall RULER accuracy compared with 89.52 for dense attention, and 0.376 on LongBench-v2 compared with 0.394 for dense attention. InfiniteBench and Video-MME extend the quality evaluation. Supplementary A100 experiments measure actual retention under the original quality configurations and examine matched-density accuracy, selection mechanisms, parameter sensitivity, and memory use.

Our contributions are:
\begin{itemize}
    \item An empirical analysis of mean dilution that links centroid underestimation to key-block dispersion and motivates radius-conditioned rescue.
    \item A radius-adaptive dual-branch selector that combines base relevance and selective rescue through separate thresholds and a mask union, integrated with regular block-sparse prefill execution.
    \item An evaluation of long-context acceleration and quality across dense, MoE, and multimodal models, supported by measured-density comparisons and analyses of the selector, its parameters, and its memory costs.
\end{itemize}

\section{Background and Motivation}
\label{sec:background}

\subsection{Long-context prefill attention}
\label{subsec:prefill_attention}

Given query, key, and value matrices $Q,K,V \in \mathbb{R}^{N \times d}$, causal self-attention computes
\begin{equation}
    O_i = \sum_{j \leq i} \frac{\exp(q_i^\top k_j / \sqrt{d})}{\sum_{\ell \leq i}\exp(q_i^\top k_\ell / \sqrt{d})} v_j.
\end{equation}
During prefill, this operation is applied to all prompt tokens before generation begins. The cost is quadratic in the prompt length $N$, which becomes expensive for 64K--128K contexts and beyond. Unlike decode attention, where each new token attends to a fixed prefix, prefill attention must process a large triangular attention matrix in a single latency-critical stage. This makes prefill acceleration complementary to KV-cache optimization and decode-time sparse attention.

Block-sparse prefill reduces this cost by partitioning the attention matrix into query and key blocks of size $B$. For a query block $i$, the algorithm selects a subset of key blocks $\mathcal{S}_i$ and executes dense attention only inside the selected block pairs. This design is attractive because block-level sparsity preserves coalesced memory access and maps naturally to GPU kernels. The main challenge is how to select $\mathcal{S}_i$ cheaply without discarding blocks that contain important evidence.

\subsection{Pattern-based and block-scoring approaches}
\label{subsec:existing_prefill}

Existing sparse-prefill systems often rely on one of two strategies. Pattern-based methods first infer an attention structure from probe queries or partial attention computation, then extend that structure to the full prefill stage. This works well when attention maps exhibit regular patterns, such as local windows, vertical retrieval columns, or slash-like long-range dependencies. However, the pattern may vary substantially across layers, heads, tasks, and positions, especially in heterogeneous long-context prompts.

Block-scoring methods instead summarize each key block and score it against query blocks. A common summary is the key-block centroid, which allows the algorithm to cheaply approximate query-key relevance at the block level. This design is simple and efficient, but it implicitly assumes that the centroid is representative of the tokens inside the block. As we discuss next, this assumption can fail in exactly the scenarios where long-context reasoning is most sensitive.

\subsection{Mean dilution and empirical observation}
\label{subsec:mean_dilution}

Consider a key block $K_b = \{k_1,\ldots,k_B\}$ with centroid
\begin{equation}
    c_b = \frac{1}{B}\sum_{t=1}^{B} k_t.
\end{equation}
A mean-only block selector estimates the relevance of the block using a score such as $q^\top c_b$. This can be misleading when a block contains a small number of highly relevant tokens and many irrelevant tokens. If $k^\star$ is a critical token but the remaining keys point in unrelated directions, then $c_b$ may be far from $k^\star$. The block may therefore receive a low mean score even though $q^\top k^\star$ is large. We call this failure mode \emph{mean dilution}.

\begin{figure*}[t]
\centering
\begin{subfigure}[t]{0.32\textwidth}
    \centering
    \includegraphics[width=\linewidth]{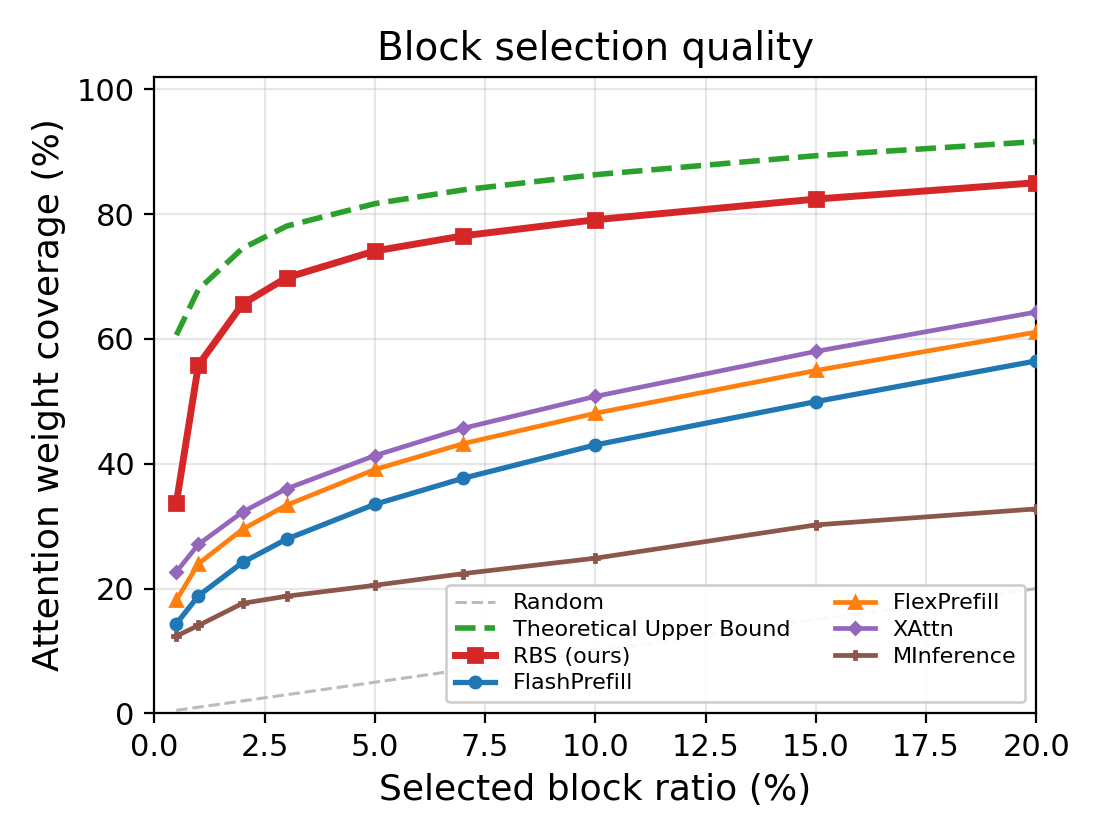}
    \caption{Block selection quality under varying budgets.}
    \label{fig:observations_a}
\end{subfigure}\hfill
\begin{subfigure}[t]{0.32\textwidth}
    \centering
    \includegraphics[width=\linewidth]{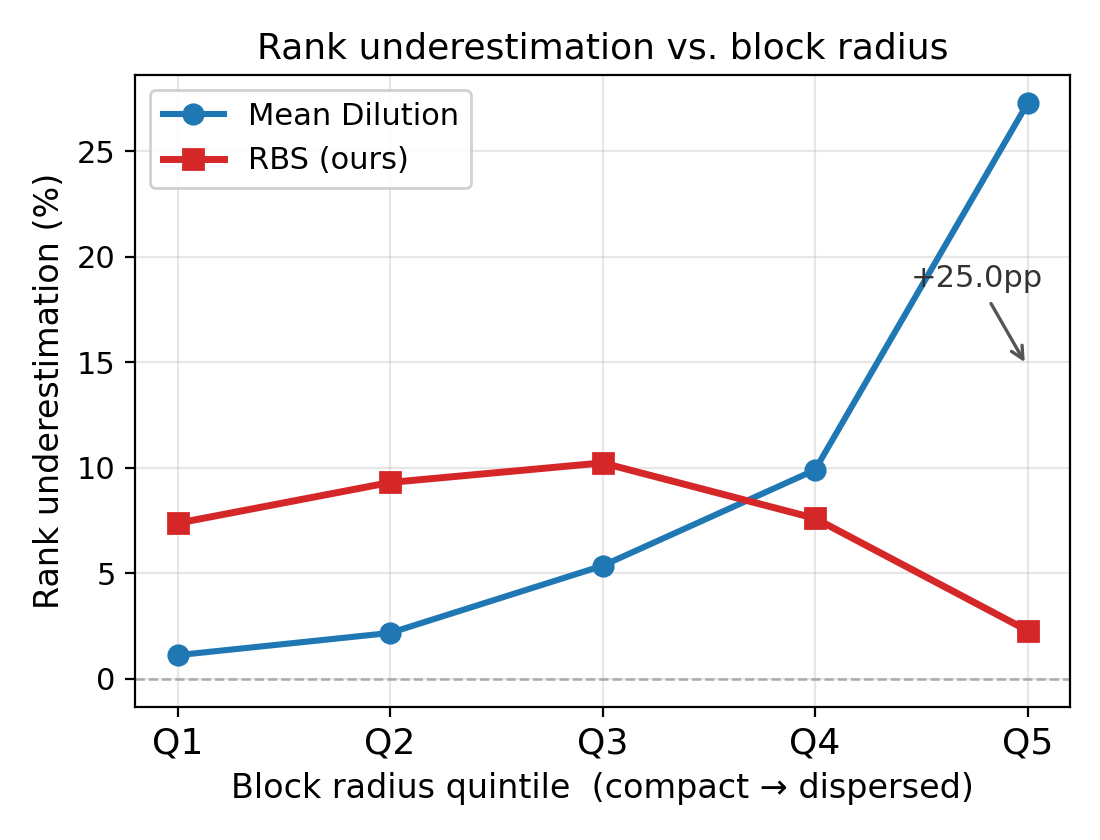}
    \caption{Mean-only scoring underestimates dispersed blocks.}
    \label{fig:observations_b}
\end{subfigure}\hfill
\begin{subfigure}[t]{0.32\textwidth}
    \centering
    \includegraphics[width=\linewidth]{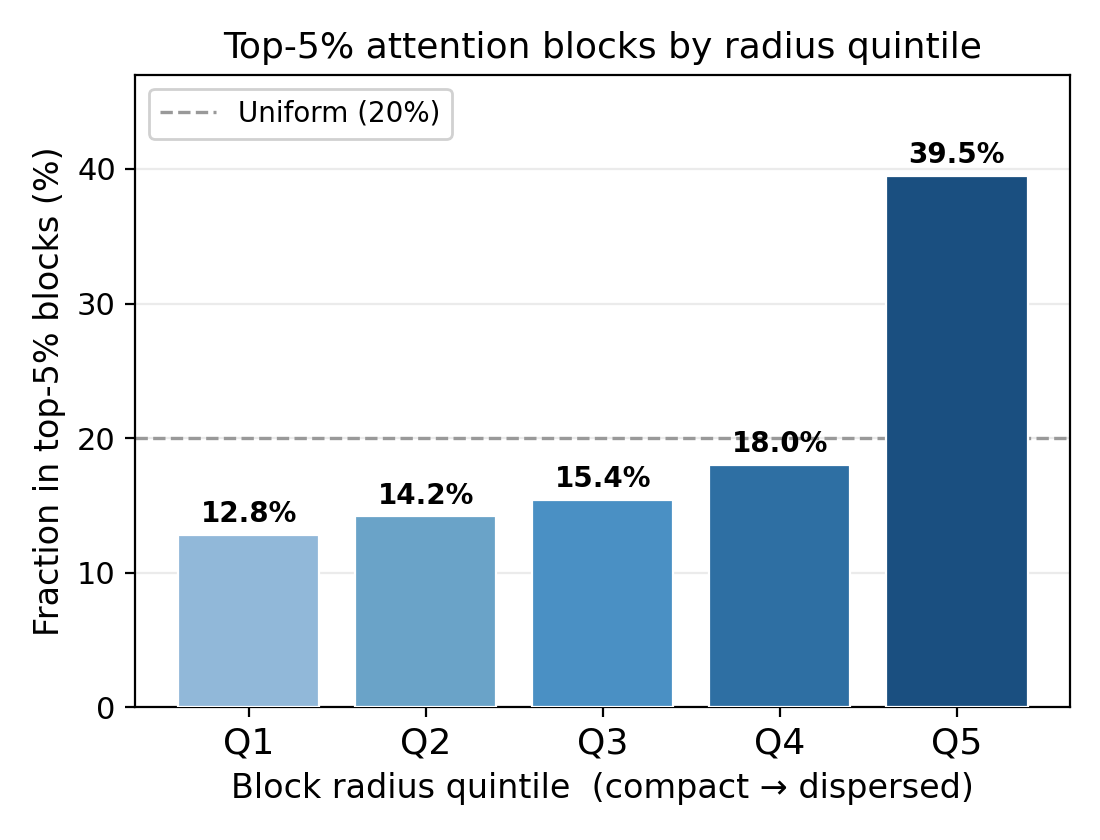}
    \caption{Q5 contains 39.5\% of top-5\% blocks.}
    \label{fig:observations_c}
\end{subfigure}
\caption{Empirical observations motivating \method. \textbf{(a)} Attention-weight coverage at a given selected-block ratio is higher for RBS than for the displayed sparse-prefill baselines. The dashed green curve denotes the reference ceiling for this coverage diagnostic, distinct from the per-token logit bound in Eq.~\eqref{eq:full_l2_bound}. \textbf{(b)} Blocks are grouped by radius quintile from compact (Q1) to dispersed (Q5). Mean-only rank underestimation increases with radius; RBS reduces underestimation in Q4 and Q5, with a 25-percentage-point difference in Q5. \textbf{(c)} Of the top-5\% attention blocks, 39.5\% fall in the highest-radius quintile, compared with a uniform reference of 20\%.}
\label{fig:observations}
\end{figure*}

Figure~\ref{fig:observations} provides empirical evidence for this motivation. Figure~\ref{fig:observations_a} shows that, under the same selected-block ratio, \method{} captures substantially more oracle attention mass than existing sparse-prefill baselines, especially in the high-sparsity regime. Figure~\ref{fig:observations_b} groups blocks by radius quintile, from compact blocks (Q1) to highly dispersed blocks (Q5), and measures rank underestimation. Mean-only rank underestimation grows with radius. \method{} reduces it in Q4 and Q5, with a 25-percentage-point difference in Q5; it does not reduce this diagnostic in the lower-radius quintiles. Figure~\ref{fig:observations_c} shows why this matters: the highest-radius quintile Q5 contains 39.5\% of top-5\% attention blocks, compared with 20\% under a uniform distribution. Together, these observations motivate a selector that keeps the efficient centroid path for compact blocks, but allocates a dedicated rescue path to high-radius blocks where mean dilution is most severe.

Mean dilution is particularly damaging in long-context evaluation. Many long-context benchmarks stress retrieval, multi-hop reasoning, and evidence aggregation, where the answer may depend on a small span buried inside a long prompt. Pruning the block containing that span removes its key and value tokens from the affected attention operation and can lead to a downstream error. This motivates a block selector that remains cheap, but is more conservative when a block contains large internal variation.

\subsection{Geometric screening and selective rescue}
\label{subsec:geometric_screening}

A useful geometric objective is the largest token logit in a key block, $s_b(q)=\max_{k\in K_b}q^\top k$. With centroid $c_b$ and radius $r_b=\max_{k\in K_b}\|k-c_b\|_2$, Cauchy--Schwarz gives
\begin{equation}
    s_b(q)\leq q^\top c_b+\|q\|_2r_b
    =U_b^{\mathrm{L2}}(q).
    \label{eq:full_l2_bound}
\end{equation}
This bound explains why the radius is useful: it limits how much an individual key can exceed the centroid score. A related construction underlies Quest~\citep{tang2024quest}, which stores coordinate-wise minima and maxima for a KV-cache page. Its query-dependent page score is the upper bound of an axis-aligned box. The box and the centroid-centered ball capture different aspects of key geometry; neither bound is uniformly tighter than the other (Appendix~\ref{app:geometry}).

A conservative bound alone does not determine an effective sparse selection policy. A large radius increases the full L2 bound even when the query is poorly aligned with the within-block deviations. Under a limited computation budget, such blocks can displace other useful candidates. \method{} therefore uses the radius as a signal of possible mean dilution, modulates its contribution using the current radius distribution, and retains a separately thresholded centroid branch. The resulting selector balances ordinary relevance with selective rescue. Its quality is an empirical question about which blocks are retained, rather than a consequence of one geometric bound dominating another.

\section{Method}
\label{sec:method}

\method{} accelerates long-context prefill by selecting key blocks through a radius-adaptive dual-branch score and then executing block-sparse attention over the selected blocks. The design is guided by three requirements: (i) block scoring should be much cheaper than dense attention, (ii) it should reduce false-negative pruning caused by mean dilution, and (iii) it should preserve the regular block-sparse execution pattern needed for high GPU efficiency.

\subsection{Block statistics and radius-adaptive coefficient}
\label{subsec:block_stats}

We partition the sequence into blocks of size $B$. For a key block $K_b = \{k_{b,1},\ldots,k_{b,B}\}$, \method{} computes its centroid
\begin{equation}
    c_b = \frac{1}{|K_b|}\sum_{k \in K_b} k,
\end{equation}
and its maximum radius
\begin{equation}
    r_b = \max_{k \in K_b} \|k-c_b\|_2.
\end{equation}
The centroid captures the average direction of the block, while the radius captures the largest token-level deviation from that centroid. The radius is informative, but applying the full radius term to every block can be overly conservative. We therefore convert the radius into a block-wise rescue coefficient.

For each layer and KV head, let $r_{\mathrm{low}}$ and $r_{\mathrm{high}}$ be the median and 90-th percentile of the block radii:
\begin{equation}
    r_{\mathrm{low}} = \operatorname{quantile}(r,0.5), \qquad
    r_{\mathrm{high}} = \operatorname{quantile}(r,0.9).
\end{equation}
For $r_{\mathrm{high}}>r_{\mathrm{low}}$, we define
\begin{equation}
    \beta_b = \operatorname{clamp}\left(\frac{r_b-r_{\mathrm{low}}}{r_{\mathrm{high}}-r_{\mathrm{low}}},0,1\right).
    \label{eq:beta}
\end{equation}
Compact blocks with radius below the median receive little or no rescue weight, while highly dispersed blocks near the top radius quantile receive the full rescue term. This normalization makes the method adaptive to the radius distribution of the current layer, head, and prompt.
The quantiles are computed from the prompt's block statistics; they do not require optimizing thresholds separately for each prompt.

\subsection{Dual-branch block scoring and selection}
\label{subsec:dual_scoring}

\method{} uses two complementary scoring branches. The base branch is a centroid score:
\begin{equation}
    \ell_{\mathrm{base}}(q,b) = q^\top c_b.
    \label{eq:base_score}
\end{equation}
This branch efficiently captures blocks whose centroid is a reliable summary. The rescue branch augments the centroid score with a radius-adaptive term derived from Eq.~\eqref{eq:full_l2_bound}:
\begin{equation}
    \ell_{\mathrm{rescue}}(q,b) = q^\top c_b + \|q\|_2 r_b \beta_b.
    \label{eq:rescue_score}
\end{equation}
For $\beta_b=1$, the rescue logit equals the full L2 upper bound. For $\beta_b<1$, it is a risk score and need not upper-bound every token logit in the block. This attenuation is deliberate: compact blocks receive little radius uplift, while dispersed blocks receive more opportunity for rescue. Appendix~\ref{app:geometry} gives the bound and its precise relationship to the attenuated score.

\paragraph{Query-block aggregation.}
Sparse prefill selects key blocks at the \emph{query-block} level rather than per query token. Let $Q_i$ denote query block $i$, and let $\mathcal{C}_i$ contain its causal candidate key blocks. For each branch $z\in\{\mathrm{base},\mathrm{rescue}\}$, the block score is proportional to the sum of exponentiated token logits. One numerically stable expression is
\begin{align}
    m^z_i &= \max_{b\in\mathcal{C}_i}\max_{q_t\in Q_i}\ell_z(q_t,b)/\sqrt{d}, \\
    S^z_{i,b} &= \sum_{q_t\in Q_i}\exp\left(\ell_z(q_t,b)/\sqrt{d}-m^z_i\right).
    \label{eq:block_aggregation}
\end{align}
The shift is shared across candidate key blocks within a branch and query block, so it cancels in relative-score comparisons. An equivalent implementation can use local log-sum-exp reductions and then restore their relative scales.

\paragraph{Independent thresholds and mask union.}
Following the relative-threshold selection strategy used in prior block-sparse prefill systems~\citep{fan2026flashprefill}, we apply separate thresholds to the two branches:
\begin{align}
    \mathcal{M}^{\mathrm{base}}_i &= \left\{b\in\mathcal{C}_i: S^{\mathrm{base}}_{i,b} \geq \alpha_{\mathrm{base}} \max_{b'\in\mathcal{C}_i} S^{\mathrm{base}}_{i,b'}\right\}, \\
    \mathcal{M}^{\mathrm{rescue}}_i &= \left\{b\in\mathcal{C}_i: S^{\mathrm{rescue}}_{i,b} \geq \alpha_{\mathrm{rescue}} \max_{b'\in\mathcal{C}_i} S^{\mathrm{rescue}}_{i,b'}\right\}.
    \label{eq:branch_thresholds}
\end{align}
The final selected set is the union
\begin{equation}
    \mathcal{M}^{\mathrm{final}}_i = \mathcal{M}^{\mathrm{base}}_i \cup \mathcal{M}^{\mathrm{rescue}}_i \cup \mathcal{M}^{\mathrm{sink/window/last}}_i.
\end{equation}
Here $\mathcal{M}^{\mathrm{sink/window/last}}_i$ denotes forced preservation of causal attention-sink, local sliding-window, and recent blocks~\citep{xiao2024efficient,beltagy2020longformer}. Forced blocks and overlap between branches are included once in the measured density of the final mask.

The two thresholds control selection aggressiveness independently; they are not fixed Top-$K$ budgets. For fixed thresholds, the number of retained blocks can vary with the prompt, layer, and head. We use \emph{budget allocation} to refer to this control over the base and rescue paths, and report actual density when comparing operating points. A smaller threshold admits more blocks within that branch.

Keeping both branches matters even though the unscaled rescue logits are at least as large as the base logits. Radius uplift changes scores by different amounts across key blocks and can also increase the maximum used by the rescue threshold. Consequently, rescue-only selection need not contain all blocks selected by the base branch. Taking the union preserves the original relevance path while admitting additional radius-supported candidates. This separation, together with the adaptive $\beta_b$, distinguishes \method{} from using the full L2 bound as a single ranking score.

\subsection{Sparse execution and complexity}
\label{subsec:sparse_execution}

After block selection, \method{} executes attention over the selected block pairs using a block-sparse FlashAttention kernel~\citep{dao2022flashattention,dao2023flashattention2}. The overall procedure is summarized in Algorithm~\ref{alg:rbs_attention}. Retained tiles use token-level dot products, causal masking, and softmax normalization over the retained causal keys, up to the chosen arithmetic precision. The approximation is the omission of unselected blocks; the output is not claimed to equal dense attention.

\begin{algorithm}[t]
\caption{\method{} sparse prefill}
\label{alg:rbs_attention}
\DontPrintSemicolon
\KwIn{Queries $Q$, keys $K$, values $V$, block size $B$, thresholds $\alpha_{\mathrm{base}},\alpha_{\mathrm{rescue}}$, sink/window/recent rules.}
\KwOut{Attention output $O$.}
Partition $Q,K,V$ into blocks.\;
\ForEach{key block $K_b$}{
    $c_b \leftarrow \frac{1}{|K_b|}\sum_{k\in K_b}k$\;
    $r_b \leftarrow \max_{k\in K_b}\|k-c_b\|_2$\;
}
Compute $r_{\mathrm{low}}=\operatorname{quantile}(r,0.5)$ and $r_{\mathrm{high}}=\operatorname{quantile}(r,0.9)$.\;
\ForEach{key block $K_b$}{
    Compute $\beta_b$ from Eq.~\eqref{eq:beta}\;
}
\ForEach{query block $Q_i$}{
    \ForEach{causal key block $K_b$}{
        Compute $\ell_{\mathrm{base}}(q_t,b)=q_t^\top c_b$ for $q_t\in Q_i$\;
        Compute $\ell_{\mathrm{rescue}}(q_t,b)=q_t^\top c_b+\|q_t\|_2r_b\beta_b$ for $q_t\in Q_i$\;
    }
    Aggregate logits into $S^{\mathrm{base}}_{i,b}$ and $S^{\mathrm{rescue}}_{i,b}$ with comparable scales across candidate blocks (Eq.~\eqref{eq:block_aggregation})\;
    $\mathcal{M}^{\mathrm{base}}_i \leftarrow \{b:S^{\mathrm{base}}_{i,b}\geq \alpha_{\mathrm{base}}\max_{b'}S^{\mathrm{base}}_{i,b'}\}$\;
    $\mathcal{M}^{\mathrm{rescue}}_i \leftarrow \{b:S^{\mathrm{rescue}}_{i,b}\geq \alpha_{\mathrm{rescue}}\max_{b'}S^{\mathrm{rescue}}_{i,b'}\}$\;
    $\mathcal{M}^{\mathrm{final}}_i \leftarrow \mathcal{M}^{\mathrm{base}}_i\cup\mathcal{M}^{\mathrm{rescue}}_i\cup\mathcal{M}^{\mathrm{sink/window/last}}_i$\;
}
$O \leftarrow \mathrm{BlockSparseFlashAttention}(Q,K,V,\{\mathcal{M}^{\mathrm{final}}_i\})$\;
\Return{$O$}\;
\end{algorithm}

Let $M=\lceil N/B\rceil$ denote the number of blocks. Centroid and radius computation costs $O(Nd)$, followed by quantile computation over $M$ radii for each layer and KV head. With the token-query aggregation in Eq.~\eqref{eq:block_aggregation}, scoring all queries against key centroids costs $O(NMd)=O(N^2d/B)$, ignoring causal constant factors. The two branches share the centroid dot products; query norms cost $O(Nd)$ and the radius uplift and token aggregation add $O(NM)$ scalar operations. Relative thresholding and mask union operate on $O(M^2)$ block scores.

Let $K_{\mathrm{avg}}$ be the average number of selected key blocks per query block. Sparse attention costs $O(MK_{\mathrm{avg}}B^2d)$. Thus, the method reduces the expensive token-to-token computation when $K_{\mathrm{avg}}\ll M$, while incurring an explicit selector cost. It does not remove quadratic dependence on $N$ in the worst case.

Persistent key summaries require $O(Md)$ storage per KV head: a centroid vector and two scalars, $r_b$ and $\beta_b$, per key block. The selected block-index list uses $O(MK_{\mathrm{avg}})$ entries. Temporary selector workspace is separate from these summaries: materialized block scores or masks use $O(M^2)$ space, and materializing all token-to-centroid logits uses $O(NM)$ space; tiled reductions can reduce the latter. Consequently, summary size alone does not characterize peak GPU memory. We report measured workspace and end-to-end peak memory in the supplementary system analysis.

\section{Experimental Setup}
\label{sec:experiments}

\subsection{Models and evaluation protocols}
\label{subsec:models_hardware}

Our primary evaluation uses NVIDIA H100 GPUs and Qwen3-30B-A3B-Instruct-2507-FP8~\citep{yang2025qwen3} for standalone and vLLM system measurements~\citep{kwon2023efficient}. We additionally evaluate the dense model Qwen3-32B to test whether the quality results extend beyond mixture-of-experts (MoE) architectures. For multimodal evaluation, we use Qwen3-VL-30B-A3B-Thinking-FP8~\citep{qwen2025qwen3vl}. Unless otherwise noted, the serving stack uses vLLM 0.10.0 and FlashAttention 2.8.3. Each speedup is normalized to dense execution with the same model, hardware, context length, and serving stack.

We supplement these primary experiments with controlled accuracy and system measurements on A100 GPUs. In this protocol, Qwen3-30B-A3B-Instruct-2507-FP8 uses tensor parallelism of two (TP=2), and Qwen3-32B uses TP=4. The FP8 checkpoint loader used in our stack imposes a shard-size divisibility constraint that prevents TP=4 for the MoE model; all methods for that model therefore use the same two-GPU configuration. The A100 experiments examine actual-density matching, selector alternatives, sensitivity, calibration, throughput, and memory. They support the interpretation of the primary results; their latency values are not interchangeable with the H100 scaling measurements.

\subsection{Baselines and selector comparisons}
\label{subsec:baselines}

The primary sparse-prefill baselines are FlashPrefill~\citep{fan2026flashprefill}, FlexPrefill~\citep{lai2025flexprefill}, MInference~\citep{jiang2024minference}, and XAttn~\citep{xu2025xattention}, together with dense attention. FlashPrefill is a close comparison because it also uses block-sparse prefill and relative-threshold selection, with mean-based block scoring. FlexPrefill, MInference, and XAttn provide alternative selection and execution strategies. The InfiniteBench comparison includes Dense, FlashPrefill, XAttn, and \method{}.

To examine the role of radius-adaptive selection, we also compare centroid-only scoring, the full L2 bound, and a Quest-style selector~\citep{tang2024quest} in a shared prefill backend. The Quest-style adaptation uses per-dimension key minima/maxima and query-dependent block selection. This experiment compares selection policies under the same prefill execution conditions; it does not measure the official Quest decoding system. The full-L2 variant uses the unattenuated radius contribution, whereas \method{} uses adaptive radius attenuation and a separately thresholded base/rescue union. The controlled selector comparison uses the same model, backend, and block size.

\subsection{Benchmarks and quality metrics}
\label{subsec:benchmarks_metrics}

RULER~\citep{hsieh2024ruler} and LongBench-v2~\citep{bai2025longbenchv2} are the main quality benchmarks. RULER tests long-context retrieval and reasoning across increasing context lengths. LongBench-v2 covers short, medium, and long tasks; the primary results use 503 samples for each model. We additionally report representative InfiniteBench~\citep{zhang2024infinitebench} tasks and Video-MME~\citep{fu2024videomme}. The Video-MME evaluation uses 1,000 samples and 32 frames per video.

The main tables report benchmark scores for the default and budget-sweep configurations. Additional selector diagnostics use the first ten examples from each of the 13 RULER-128K tasks, totaling 130 examples. This controlled subset has a different evaluation scope from the primary RULER benchmark, so its absolute scores are interpreted within the corresponding table. RULER and Video-MME scores are reported as percentages; LongBench-v2 and InfiniteBench scores use the benchmark's unit-interval scale.

\subsection{Actual density and hyperparameters}
\label{subsec:hyperparams}

We define actual density as
\begin{equation}
\rho = \frac{\#\{\text{selected valid causal block pairs}\}}
{\#\{\text{all valid causal block pairs}\}}.
\label{eq:actual_density}
\end{equation}
The numerator counts the final pairs entering the sparse-attention kernel, including the base/rescue union and any preserved sink, local-window, or recent blocks. We first compute density for each sample, then report its mean and, where available, its median (p50) and 90th percentile (p90). A target budget is a calibration target rather than a guarantee that different methods or prompts retain exactly the same fraction. We therefore report measured density beside quality scores and distinguish the fixed system-prompt density from the accuracy-set density.

Unless stated otherwise, RBS and the controlled selector comparisons use query and key block size $B=128$. Other baselines retain their native selection layouts. The radius-adaptive coefficient uses the median and 90th-percentile block radii, $r_{\mathrm{low}}=\operatorname{quantile}(r,0.5)$ and $r_{\mathrm{high}}=\operatorname{quantile}(r,0.9)$. The default thresholds are $\alpha_{\mathrm{base}}=0.22$ and $\alpha_{\mathrm{rescue}}=0.18$. FlashPrefill uses its tuned $\alpha=0.12$ setting and XAttn uses $\mathrm{stride}=16$ in the primary quality experiments. Budget sweeps and sensitivity experiments recalibrate their selection parameters as specified in the respective tables. Thresholds are fixed after calibration; the prompt-dependent radius normalization is computed directly at runtime without a per-prompt parameter search.

For the primary LongBench-v2 configurations, auxiliary density measurements use a fixed-seed selector-only sample of 20 examples from each length category. For the primary RULER configurations, density is measured on one example per task at each context length. These measurements estimate the budgets of the original configurations; they are not full-dataset density measurements. The accuracy scores themselves retain the full primary evaluation.

\subsection{System metrics}
\label{subsec:system_metrics}

The primary H100 measurements span 16K--256K context and report (i) standalone prefill-attention speedup, (ii) vLLM prefill-attention speedup, and (iii) end-to-end time-to-first-token (TTFT) speedup. TTFT measures elapsed time from request arrival until the first output token is available. These metrics respectively characterize standalone attention execution, the attention path in the serving stack, and user-visible latency.

The supplementary A100 measurements additionally report selector time, sparse-attention time, prompt throughput, and memory. Selector time includes block statistics, normalization, score computation, thresholding or Top-K, mask union when applicable, and index compaction. Prompt throughput is the number of processed prompt tokens divided by batch wall-clock time, normalized to Dense at the same context and batch size. Peak allocated memory is the maximum of \texttt{torch.cuda.max\_memory\_allocated} over TP ranks. We report persistent block statistics and temporary workspace separately because neither alone characterizes the full-model memory peak.

\section{Results}
\label{sec:results}

\subsection{Long-context acceleration}
\label{subsec:speed_results}

\begin{figure}[htbp]
    \centering
    \includegraphics[width=\textwidth]{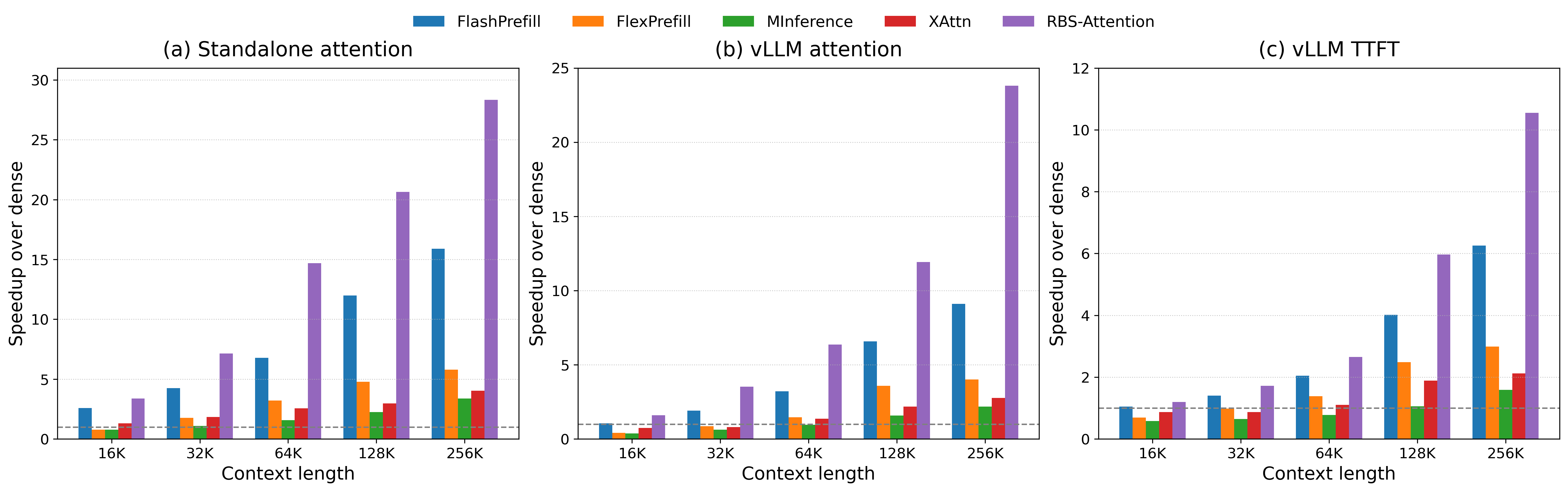}
    \caption{long-context speedup measurements on Qwen3-30B-A3B-Instruct-2507-FP8 using H100 GPUs. \textbf{(a)} Standalone prefill-attention speedup. \textbf{(b)} vLLM prefill-attention speedup. \textbf{(c)} End-to-end vLLM TTFT speedup. Every speedup is normalized to Dense at the same context length and under the same execution setup. These are the default operating configurations; supplementary budget-calibrated comparisons with measured densities are reported in Appendix~\ref{app:controlled_system}.}
    \label{fig:speed_3panel}
\end{figure}

Figure~\ref{fig:speed_3panel} reports the H100 system measurements on Qwen3-30B-A3B-Instruct-2507-FP8. At 128K context, \method{} achieves 20.65$\times$ standalone prefill-attention speedup, compared with 11.98$\times$ for FlashPrefill, 4.78$\times$ for FlexPrefill, 2.25$\times$ for MInference, and 2.97$\times$ for XAttn. Within vLLM, \method{} reaches 11.92$\times$ prefill-attention speedup and 5.97$\times$ end-to-end TTFT speedup at 128K.

The gains grow in the long-context regime, where the attention work avoided by block selection becomes large enough to amortize selector overhead. At shorter contexts, some sparse methods approach or fall below dense speed. This behavior motivates evaluating quality and retained density alongside speed: an effective prefill selector must avoid both excessive selection overhead and the loss of task-relevant blocks. The following quality experiments evaluate this trade-off, and the supplementary controlled experiments further examine differences in actual retained budget.

\FloatBarrier

\subsection{Budget--accuracy trade-off on RULER 128K}
\label{subsec:budget_accuracy}

Table~\ref{tab:ruler_budget} reports the budget sweep on RULER 128K with Qwen3-30B-A3B-Instruct-2507-FP8. Dense attention obtains 89.69\% accuracy. \method{} has the highest sparse accuracy at each of the four nominal target budgets, reaching 80.28\% at 5.34\% actual density and 88.36\% at 11.03\% actual density. The measured densities differ across methods, so the nominal-budget table alone does not establish an exactly matched-budget comparison.

\begin{table}[htbp]
\centering
\caption{RULER-128K budget sweep on Qwen3-30B-A3B-Instruct-2507-FP8. Each cell reports \textbf{actual density / accuracy}, both in percent. Dense obtains 89.69\% accuracy at 100\% density. Bold denotes the highest sparse accuracy within each nominal target row; actual densities are reported explicitly because the targets are not exact matches.}
\label{tab:ruler_budget}
\small
\setlength{\tabcolsep}{4pt}
\begin{tabular}{@{}lccccc@{}}
\toprule
Target & FlashPrefill & XAttn & FlexPrefill & MInference & RBS-Attention \\
\midrule
1\% & 1.77 / 56.65 & 1.02 / 44.51 & 1.17 / 31.64 & 1.01 / 50.83 & 1.80 / \textbf{60.36} \\
2\% & 1.91 / 63.32 & 1.71 / 50.28 & 2.25 / 41.41 & 1.82 / 68.58 & 2.03 / \textbf{70.51} \\
5\% & 4.69 / 75.36 & 3.81 / 60.47 & 3.59 / 50.72 & 5.90 / 76.41 & 5.34 / \textbf{80.28} \\
10\% & 10.19 / 85.01 & 7.63 / 71.01 & 6.69 / 65.87 & 11.45 / 87.18 & 11.03 / \textbf{88.36} \\
\bottomrule
\end{tabular}
\end{table}

To check whether the observed advantage persists when budgets are aligned more tightly, we recalibrate baselines to the RBS density anchors (Appendix~\ref{app:matched_accuracy}). At the 5.34\% anchor, with every method within 0.10 percentage point of the target, RBS obtains 80.28 accuracy, compared with 76.87 for MInference, 74.73 for FlashPrefill, 66.77 for XAttn, and 58.56 for FlexPrefill. This additional comparison supports the observation that the base/rescue allocation preserves useful blocks under a limited compute budget.

\FloatBarrier

\subsection{Main quality results: RULER and LongBench-v2}
\label{subsec:main_quality}

Table~\ref{tab:longbench_dual} reports LongBench-v2 quality and measured density estimates for the evaluated configurations. On Qwen3-32B, RBS obtains 0.376, matching the best sparse total score, with 9.719\% estimated mean density. On Qwen3-30B-A3B-Instruct-2507-FP8, RBS obtains 0.370 with 7.912\% estimated mean density, compared with Dense's 0.388. The density estimates make the operating points more informative: for example, XAttn matches RBS's dense-model total score while retaining an estimated 29.197\% of block pairs, and MInference retains more than half of the pairs on average on both models.

\begin{table}[htbp]
\centering
\caption{LongBench-v2 quality with supplementary density measurements. Accuracy uses 503 samples per model. Density (mean/p50/p90, in percent) is estimated using a fixed-seed selector-only sample of 20 short, 20 medium, and 20 long examples under the original configurations. It is not a full-503-sample density measurement. Bold denotes the best sparse accuracy in each column.}
\label{tab:longbench_dual}
\small
\setlength{\tabcolsep}{4pt}
\begin{tabular}{@{}lccccrrr@{}}
\toprule
& \multicolumn{4}{c}{Accuracy} & \multicolumn{3}{c}{Estimated density (\%)} \\
\cmidrule(lr){2-5}\cmidrule(lr){6-8}
Method & Short & Medium & Long & Total & Mean & p50 & p90 \\
\midrule
\multicolumn{8}{c}{Qwen3-32B} \\
\midrule
Dense & 0.450 & 0.367 & 0.380 & 0.394 & 100 & 100 & 100 \\
FlashPrefill & 0.378 & 0.349 & 0.361 & 0.362 & 14.149 & 9.197 & 28.047 \\
XAttn & \textbf{0.439} & 0.340 & \textbf{0.370} & \textbf{0.376} & 29.197 & 26.711 & 43.858 \\
FlexPrefill & 0.422 & 0.293 & 0.352 & 0.352 & 11.397 & 10.203 & 16.260 \\
MInference & 0.417 & 0.335 & 0.315 & 0.360 & 55.763 & 44.204 & 87.840 \\
RBS-Attention & 0.411 & \textbf{0.353} & 0.361 & \textbf{0.376} & 9.719 & 5.515 & 20.172 \\
\midrule
\multicolumn{8}{c}{Qwen3-30B-A3B-Instruct-2507-FP8} \\
\midrule
Dense & 0.422 & 0.358 & 0.388 & 0.388 & 100 & 100 & 100 \\
FlashPrefill & 0.400 & 0.340 & 0.315 & 0.356 & 11.060 & 7.269 & 22.325 \\
XAttn & 0.383 & 0.316 & 0.352 & 0.348 & 26.129 & 24.297 & 37.965 \\
FlexPrefill & 0.328 & \textbf{0.350} & 0.343 & 0.340 & 16.544 & 15.863 & 23.124 \\
MInference & \textbf{0.406} & 0.302 & 0.352 & 0.348 & 51.851 & 40.397 & 81.776 \\
RBS-Attention & \textbf{0.406} & 0.340 & \textbf{0.370} & \textbf{0.370} & 7.912 & 4.638 & 16.367 \\
\bottomrule
\end{tabular}
\end{table}

\begin{table}[htbp]
\centering
\caption{RULER accuracy on Qwen3-32B with supplementary actual-density estimates. Each method has an accuracy row followed by an italic density row; all values are percentages. Accuracy uses the benchmark evaluation. Density is measured on one example per task at each length, so it estimates the original configuration's budget. Overall is the reported average across lengths. Bold denotes the highest sparse accuracy.}
\label{tab:ruler}
\small
\setlength{\tabcolsep}{4pt}
\begin{tabular}{@{}lrrrrrrr@{}}
\toprule
Method & 4K & 8K & 16K & 32K & 64K & 128K & Overall \\
\midrule
Dense & 93.30 & 91.90 & 92.00 & 89.70 & 89.50 & 80.70 & 89.52 \\
\textit{density} & \textit{100} & \textit{100} & \textit{100} & \textit{100} & \textit{100} & \textit{100} & \textit{100} \\
\midrule
FlashPrefill & 91.00 & 89.20 & 88.80 & 81.50 & 81.80 & 68.10 & 83.40 \\
\textit{density} & \textit{81.091} & \textit{59.781} & \textit{40.970} & \textit{26.754} & \textit{16.872} & \textit{10.075} & \textit{39.257} \\
XAttn & 92.40 & \textbf{91.30} & 90.90 & 86.80 & 87.20 & 79.20 & 87.97 \\
\textit{density} & \textit{69.518} & \textit{59.482} & \textit{51.776} & \textit{43.645} & \textit{36.201} & \textit{29.196} & \textit{48.303} \\
FlexPrefill & 72.30 & 75.10 & 78.30 & 79.00 & 77.70 & 73.60 & 76.00 \\
\textit{density} & \textit{30.000} & \textit{23.689} & \textit{20.140} & \textit{16.915} & \textit{14.999} & \textit{11.719} & \textit{19.577} \\
MInference & \textbf{92.61} & 91.24 & 90.41 & 87.59 & 88.96 & \textbf{80.96} & 88.63 \\
\textit{density} & \textit{100.000} & \textit{99.990} & \textit{97.353} & \textit{86.940} & \textit{70.251} & \textit{47.968} & \textit{83.750} \\
RBS-Attention & 91.10 & 89.90 & \textbf{91.20} & \textbf{89.50} & \textbf{89.60} & 80.57 & \textbf{88.65} \\
\textit{density} & \textit{73.030} & \textit{48.465} & \textit{30.185} & \textit{18.224} & \textit{10.720} & \textit{6.028} & \textit{31.109} \\
\bottomrule
\end{tabular}
\end{table}

Table~\ref{tab:ruler} reports the RULER results on Qwen3-32B. RBS scores 88.65 overall, compared with 89.52 for Dense and 88.63 for MInference. At 128K, RBS reaches 80.57 versus Dense's 80.70 while retaining an estimated 6.028\% of causal block pairs. This result also demonstrates that the method's quality preservation extends to a dense architecture. The density estimates decrease as context grows for these fixed-threshold configurations, highlighting why a threshold should not be interpreted as a fixed sparsity ratio.

\FloatBarrier

\subsection{Additional results: InfiniteBench and Video-MME}
\label{subsec:additional_results}

Table~\ref{tab:InfiniteBench} extends the text evaluation to representative InfiniteBench tasks. RBS obtains a macro average of 0.362, compared with 0.338 for FlashPrefill and 0.352 for XAttn. Improvements over FlashPrefill are particularly apparent for key--value retrieval and long-dialogue question answering. These results provide additional evidence that radius-adaptive selection can preserve useful context across tasks beyond RULER and LongBench-v2.

\begin{table}[htbp]
\centering
\caption{InfiniteBench results on Qwen3-32B. Task names follow the benchmark abbreviations. Scores use the unit-interval scale; the last row is the reported macro average over the listed tasks. Bold denotes the best sparse score.}
\label{tab:InfiniteBench}
\begin{tabular}{@{}lrrrr@{}}
\toprule
Task & Dense & FlashPrefill & XAttn & RBS-Attention \\
\midrule
Ret.KV & 0.244 & 0.148 & 0.194 & \textbf{0.204} \\
Ret.PassKey & 1.000 & \textbf{1.000} & \textbf{1.000} & \textbf{1.000} \\
Ret.Num & 1.000 & \textbf{1.000} & \textbf{1.000} & 0.998 \\
En.Sum & 0.289 & 0.291 & 0.272 & \textbf{0.296} \\
En.MC & 0.105 & 0.100 & \textbf{0.105} & 0.092 \\
En.QA & 0.087 & 0.075 & \textbf{0.084} & 0.083 \\
Zh.QA & 0.115 & 0.106 & 0.109 & \textbf{0.112} \\
En.Dia & 0.360 & 0.170 & 0.310 & \textbf{0.325} \\
Math.Find & 0.451 & \textbf{0.463} & 0.413 & 0.459 \\
Code.Debug & 0.066 & 0.028 & 0.030 & \textbf{0.053} \\
\midrule
Macro average & 0.372 & 0.338 & 0.352 & \textbf{0.362} \\
\bottomrule
\end{tabular}
\end{table}

Table~\ref{tab:videomme} reports the multimodal evaluation. RBS obtains 65.98 overall on Video-MME, compared with 66.16 for Dense, and has the highest total score among the sparse methods in this comparison. The long-video score is 62.1. This is supporting quality evidence for multimodal prefill; the H100 text-model speedups should not be read as measurements of this vision-language model.

\begin{table}[htbp]
\centering
\caption{Video-MME results on Qwen3-VL-30B-A3B-Thinking-FP8, using 1,000 samples and 32 frames per video. Scores are percentages. Bold denotes the best sparse score.}
\label{tab:videomme}
\begin{tabular}{@{}lrrrr@{}}
\toprule
Method & Short & Medium & Long & Total \\
\midrule
Dense & 73.6 & 62.5 & 58.6 & 66.16 \\
\midrule
FlashPrefill & 71.2 & \textbf{61.5} & 60.9 & 65.26 \\
XAttn & \textbf{72.8} & 61.0 & 61.2 & 65.76 \\
FlexPrefill & 71.0 & 57.9 & 56.3 & 62.82 \\
MInference & 71.4 & 60.5 & 59.8 & 64.72 \\
RBS-Attention & 72.7 & 61.2 & \textbf{62.1} & \textbf{65.98} \\
\bottomrule
\end{tabular}
\end{table}

\FloatBarrier

\subsection{Supporting evidence for adaptive rescue}
\label{subsec:selector_evidence}

Table~\ref{tab:selector_ablation} examines whether a full geometric bound alone reproduces the benefit of RBS. All four selectors use the same prefill backend, $B=128$, the same 130 RULER-128K examples, and actual density within 0.10 percentage point of 5.34\%. RBS has the highest point estimate at 76.67, compared with 73.63 for centroid-only, 72.41 for the unattenuated full-L2 selector, and 73.71 for Quest-style selection. This pattern is consistent with the role of adaptive base/rescue allocation: making every radius contribution maximally conservative is not necessarily the best use of a limited retained budget. However, the paired bootstrap difference intervals between RBS and each comparator include zero; this small diagnostic set does not establish statistical significance.

\begin{table}[htbp]
\centering
\caption{Supporting selector comparison on 130 RULER-128K examples with Qwen3-30B-A3B-Instruct-2507-FP8. All selectors share the prefill backend and $B=128$. Accuracy and density are percentages. Intervals are sample-bootstrap 95\% confidence intervals for each accuracy estimate, not intervals for between-method differences. Quest-style denotes a controlled prefill adaptation.}
\label{tab:selector_ablation}
\begin{tabular}{@{}lrrr@{}}
\toprule
Selector & Actual density & Accuracy & 95\% CI \\
\midrule
Centroid-only & 5.300 & 73.63 & [68.00, 81.21] \\
Full-L2 bound & 5.295 & 72.41 & [66.32, 79.79] \\
Quest-style & 5.334 & 73.71 & [66.76, 80.47] \\
RBS adaptive union & 5.350 & \textbf{76.67} & [69.01, 82.15] \\
\bottomrule
\end{tabular}
\end{table}

The supplementary experiments characterize the same design from several additional angles. The A100 system sweep reports 4.47$\times$ TTFT speedup for RBS at the 5\% target, with 4.792\% density on the system prompt (Appendix~\ref{app:controlled_system}). Block-size and cross-model threshold sweeps expose the dependence of quality and actual density on configuration (Appendix~\ref{app:sensitivity}). Calibration, runtime adaptation, and memory measurements quantify deployment costs (Appendix~\ref{app:deployment}). These measurements complement the H100 scaling and benchmark results by isolating selection behavior and quantifying deployment costs.

\FloatBarrier
\section{Related Work}
\label{sec:related_work}

\paragraph{Efficient attention kernels.}
FlashAttention and FlashAttention-2 accelerate dense attention through tiling, reduced memory traffic, and improved GPU work partitioning~\citep{dao2022flashattention,dao2023flashattention2}. \method{} combines block selection with this execution approach: it reduces the set of computed tiles and applies a FlashAttention-style kernel to the retained causal keys. Kernel acceleration and block selection address different parts of the attention cost.

\paragraph{Sparse prefill attention.}
Recent sparse-prefill systems infer attention structure from probe queries, sample the attention matrix, or construct coarse block scores~\citep{jiang2024minference,lai2025flexprefill,xu2025xattention,fan2026flashprefill}. These methods offer useful accuracy--latency trade-offs, but their selectors can respond differently to isolated evidence and heterogeneous blocks. Our focus is mean dilution in centroid summaries. \method{} retains centroid relevance while adding a radius-conditioned rescue path, with independent relative thresholds and a union of the two masks. The relative-threshold operation follows prior prefill practice~\citep{fan2026flashprefill}; the additional contribution is how radius-derived risk is incorporated into selection.

\paragraph{Query-aware KV-page selection.}
Quest~\citep{tang2024quest} maintains coordinate-wise minimum and maximum keys per KV-cache page, scores pages using the current query, and loads the Top-$K$ pages for decoding attention. \method{} shares the principle of using query-aware geometry to avoid missing a high-scoring token. It addresses prefill block pairs, however, and uses an attenuated radius score alongside a centroid branch. Its thresholds induce a variable selected density, whereas Quest uses a fixed page budget. The distinction is in the selection policy and execution stage as well as the geometric summary; we do not claim that the L2 bound is uniformly tighter than Quest's box bound. Appendix~\ref{app:geometry} makes the geometric relationship explicit, and the controlled Quest-style experiment isolates page-scoring behavior within the prefill evaluation.

\paragraph{Heterogeneity-aware sparse attention.}
SpargeAttention~\citep{zhang2025spargeattention} also recognizes that mean compression can be unreliable for heterogeneous blocks. Its first stage uses self-similarity to decide which query and key blocks can be compressed for sparse prediction; blocks failing that test are protected from first-stage pruning. Its complete operator additionally uses an online softmax-aware filter and integrates quantized attention. \method{} uses a continuous coefficient derived from key-block radius quantiles and a separately thresholded rescue branch. Thus, a dispersed block receives a larger rescue score but is not automatically retained solely because it is dispersed. These are distinct policies for handling unreliable block summaries, and our discussion concerns that algorithmic relationship rather than a comparison of the complete operators.

\section{Limitations and Discussion}
\label{sec:limitations}

\paragraph{Execution regime.}
RBS targets long-context prefill, where reducing token-level attention work can amortize selection overhead. Its benefit is smaller at short contexts. The geometric statistics could inform decode-time retrieval, but our query-block selection and system results concern prefill. Decode has a different balance of query count, KV-cache traffic, and access overhead, and requires a separate implementation and evaluation.

\paragraph{Selection and parameter dependence.}
A large radius indicates potential mean dilution, but does not establish that a block contains evidence relevant to a particular query. Thresholds control both branches, and the final density varies with model, prompt, layer, and head. The block-size and threshold studies characterize the tested operating points; broader evaluation is needed to determine how well these settings transfer. The controlled selector comparison also has a limited sample size, and its uncertainty is discussed alongside the results.

\paragraph{Evaluation and memory scope.}
The main speed measurements use H100 GPUs; the supplementary deployment measurements use A100 GPUs with their stated tensor-parallel configurations. We do not infer speedups on the dense or multimodal models from the MoE timing results. Densities accompanying the original quality tables are estimates from a smaller selector-only sample, with the sampling protocol reported beside the tables. The Quest-style experiment isolates a selection policy in our prefill backend and is not a benchmark of the complete original decode system. RBS adds little persistent radius metadata, but its dual-branch implementation also requires temporary workspace; both are reported separately from the full-model memory peak.

\section{Conclusion}
\label{sec:conclusion}

We introduced \method{}, a training-free method for sparse long-context prefill. It retains centroid-based selection while providing a separately thresholded rescue branch whose contribution adapts to the distribution of key-block radii. This design addresses mean dilution through selective use of block dispersion and preserves regular block-sparse attention execution. The primary H100 evaluation achieves 20.65$\times$ standalone prefill-attention speedup and 5.97$\times$ end-to-end TTFT speedup at 128K, alongside quality evaluation on RULER, LongBench-v2, InfiniteBench, and Video-MME. Supporting measurements clarify actual sparsity, the distinction from full-bound selection, and the costs of calibration and runtime adaptation. These results demonstrate the value of combining average relevance with radius-adaptive rescue for efficient long-context prefill.

\clearpage
\begingroup
\small
\setlength{\bibsep}{4pt plus 1pt}
\interlinepenalty=10000
\bibliography{references}
\bibliographystyle{plainnat}
\endgroup

\clearpage
\appendix
\section{Additional Accuracy and Budget Analysis}
\label{app:matched_accuracy}

The supplementary experiments examine actual retention under nominal budgets, the effect of the selection policy, and deployment behavior. Unless otherwise specified, the model in these controlled experiments is Qwen3-30B-A3B-Instruct-2507-FP8.

\subsection{Strict actual-density matching}

Table~\ref{tab:strict_density} recalibrates competing methods to the actual RBS density anchors from the primary budget sweep. Every added point is within 0.10 percentage point of its anchor. RBS has the highest accuracy among the measured methods at each anchor. At 5.34\%, the RBS advantage is 3.41 accuracy points over MInference and 5.55 points over FlashPrefill. This supports the interpretation of Table~\ref{tab:ruler_budget} at a more closely aligned retained budget. These are accuracy-only calibration points; they should not be paired with latency from a different threshold configuration.

\begin{table}[htbp]
\centering
\caption{Additional RULER-128K accuracy measurements with strict actual-density matching. Each cell gives \textbf{actual density / accuracy} in percent. A dash means that no additional point was measured at that anchor.}
\label{tab:strict_density}
\begin{tabular}{@{}lccc@{}}
\toprule
Method & 2.03\% anchor & 5.34\% anchor & 11.03\% anchor \\
\midrule
FlashPrefill & 2.038 / 67.13 & 5.300 / 74.73 & 11.035 / 79.62 \\
MInference & 1.944 / 68.40 & 5.286 / 76.87 & 10.950 / 83.44 \\
XAttn & --- & 5.311 / 66.77 & --- \\
FlexPrefill & --- & 5.352 / 58.56 & --- \\
RBS-Attention & 2.030 / \textbf{70.51} & 5.340 / \textbf{80.28} & 11.030 / \textbf{88.36} \\
\bottomrule
\end{tabular}
\end{table}

\subsection{Selector diagnostics}

The controlled selector comparison in Table~\ref{tab:selector_ablation} uses the first ten examples of each RULER-128K task, with the same model, backend, and $B=128$. The centroid-only and full-L2 selectors use selection thresholds of 0.194 and 0.022, respectively, and Quest-style selection uses its density parameter 0.0405. Because the selection parameters have different meanings across policies, the comparison is aligned using measured density.

The retrieval and multi-key breakdowns are provided in Table~\ref{tab:selector_breakdown}. Their point estimates agree with the aggregate trend, but the limited diagnostic sample warrants interpreting these differences descriptively. These accuracy experiments do not establish a universal ordering of the mathematical tightness of box and L2 bounds.

\begin{table}[htbp]
\centering
\caption{Task-group diagnostics for the same 130-example selector comparison as Table~\ref{tab:selector_ablation}. Scores are percentages.}
\label{tab:selector_breakdown}
\begin{tabular}{@{}lrr@{}}
\toprule
Selector & Retrieval accuracy & Multi-key accuracy \\
\midrule
Centroid-only & 88.75 & 90.00 \\
Full-L2 bound & 87.81 & 90.00 \\
Quest-style & 88.75 & 90.00 \\
RBS adaptive union & 90.00 & 93.33 \\
\bottomrule
\end{tabular}
\end{table}

\FloatBarrier
\section{Additional A100 System Measurements}
\label{app:controlled_system}

\subsection{Complete 128K sweep}

Table~\ref{tab:a100_sweep} reports the additional 128K system sweep on Qwen3-30B-A3B-Instruct-2507-FP8 using A100 GPUs and TP=2. The system measurements use a fixed prompt; the accuracy entries are companion evaluation results. Since the prompt distribution and selection configurations affect retention, the two measured densities are shown separately. These nominal-target companion points are distinct from the strict recalibrations in Table~\ref{tab:strict_density} and the 130-example selector ablation in Table~\ref{tab:selector_ablation}.

RBS reaches 5.29$\times$, 4.47$\times$, and 3.05$\times$ TTFT speedup at the 2\%, 5\%, and 10\% targets, respectively. The corresponding companion accuracy values are 70.51, 80.28, and 88.36. At the central target, the other measured TTFT speedups range from 2.84$\times$ to 3.58$\times$. This provides supporting system evidence at low measured density. Because actual densities and accuracy evaluation sets are reported separately, the table should not be interpreted as an exactly equal-density Pareto frontier.

\begin{table}[htbp]
\centering
\caption{Supplementary A100 128K sweep on Qwen3-30B-A3B-Instruct-2507-FP8 (TP=2). Sys./acc. are actual densities on the system prompt and companion accuracy evaluation, respectively; densities and accuracy are percentages. Selector and sparse-attention shares sum to 100\% of their reported combined path, not TTFT. Dense's 86.31 is the controlled-subset reference, not the primary full-benchmark 89.69. Companion accuracies retain their source evaluation protocols; the additional 2\%/10\% points for XAttn, FlexPrefill, and Quest-style use 130 examples.}
\label{tab:a100_sweep}
\small
\setlength{\tabcolsep}{4pt}
\begin{tabular}{@{}lrrrrrr@{}}
\toprule
Method & Target & Sys./acc. density & Selector & Sparse attn. & TTFT gain & Accuracy \\
\midrule
Dense & 100 & 100 / 100 & --- & 100.00 & 1.00$\times$ & 86.31 \\
\midrule
FlashPrefill & 2 & 1.826 / 1.910 & 23.87 & 76.13 & 4.89$\times$ & 63.32 \\
& 5 & 5.071 / 4.690 & 12.39 & 87.61 & 2.90$\times$ & 75.36 \\
& 10 & 11.147 / 10.190 & 5.99 & 94.01 & 2.04$\times$ & 85.01 \\
\midrule
MInference & 2 & 1.568 / 2.022 & 41.11 & 58.89 & 4.84$\times$ & 68.54 \\
& 5 & 4.184 / 5.056 & 26.61 & 73.39 & 3.58$\times$ & 76.09 \\
& 10 & 8.609 / 9.995 & 16.61 & 83.39 & 2.93$\times$ & 82.82 \\
\midrule
RBS-Attention & 2 & 1.830 / 2.030 & 31.68 & 68.32 & 5.29$\times$ & 70.51 \\
& 5 & 4.792 / 5.340 & 16.28 & 83.72 & 4.47$\times$ & 80.28 \\
& 10 & 10.517 / 11.030 & 7.73 & 92.27 & 3.05$\times$ & 88.36 \\
\midrule
XAttn & 2 & 1.957 / 1.971 & 72.14 & 27.86 & 5.01$\times$ & 52.62 \\
& 5 & 4.105 / 3.810 & 53.07 & 46.93 & 3.55$\times$ & 60.47 \\
& 10 & 10.588 / 10.970 & 33.38 & 66.62 & 1.74$\times$ & 76.51 \\
\midrule
FlexPrefill & 2 & 2.908 / 2.041 & 45.12 & 54.88 & 4.51$\times$ & 41.95 \\
& 5 & 5.269 / 3.590 & 29.00 & 71.00 & 3.21$\times$ & 50.72 \\
& 10 & 10.762 / 11.065 & 19.87 & 80.13 & 2.32$\times$ & 77.90 \\
\midrule
Quest-style & 2 & 1.851 / 1.972 & 52.38 & 47.62 & 4.22$\times$ & 62.32 \\
& 5 & 4.653 / 4.807 & 34.73 & 65.27 & 2.84$\times$ & 70.56 \\
& 10 & 10.953 / 11.098 & 20.64 & 79.36 & 2.06$\times$ & 78.63 \\
\bottomrule
\end{tabular}
\end{table}

\subsection{Prompt throughput}

Table~\ref{tab:a100_throughput} reports prompt throughput for the 5\% target setting. Context denotes the nominal length bucket; the Dense column gives actual total prompt tokens and wall time. Throughput gains are computed against Dense separately for every context and batch size. RBS reaches 4.47$\times$ throughput at 128K with batch size one and 3.85$\times$ at batch size four. Shorter contexts show smaller gains, consistent with the cost of computing a sparse mask and dispatching sparse kernels.

The MInference and FlexPrefill kernels used in this experiment accept a single sequence. For batch size four, we invoke them sequentially on four independent causal sequences and include dispatch costs in batch wall time. Consequently, their batched results characterize this adaptation rather than an optimized native batching implementation. No requests are concatenated into one causal sequence.

\begin{table}[htbp]
\centering
\caption{Supplementary A100 prompt throughput at the 5\% target, using Qwen3-30B-A3B-Instruct-2507-FP8 (TP=2). The upper panel provides absolute Dense measurements; the lower panel gives method tokens/s divided by Dense tokens/s for the same row. FP = FlashPrefill; MI = MInference; Flex = FlexPrefill. Quest denotes the controlled prefill adaptation.}
\label{tab:a100_throughput}
\small
\begin{tabular}{@{}rrrrr@{}}
\toprule
Context & Batch & Dense wall time (ms) & Total prompt tokens & Dense tokens/s \\
\midrule
32K & 1 & 1,920.68 & 31,286 & 16,289 \\
64K & 1 & 5,734.91 & 62,486 & 10,896 \\
128K & 1 & 19,709.12 & 125,486 & 6,367 \\
32K & 4 & 6,996.10 & 118,097 & 16,880 \\
64K & 4 & 21,265.25 & 239,377 & 11,257 \\
128K & 4 & 72,127.35 & 478,727 & 6,637 \\
\bottomrule
\end{tabular}
\par\medskip
\begin{tabular}{@{}rrcccccc@{}}
\toprule
Context & Batch & FP & MI & RBS & XAttn & Flex & Quest \\
\midrule
32K & 1 & 1.25$\times$ & 1.36$\times$ & 1.27$\times$ & 0.98$\times$ & 0.95$\times$ & 1.29$\times$ \\
64K & 1 & 2.05$\times$ & 2.24$\times$ & 2.06$\times$ & 1.49$\times$ & 1.69$\times$ & 2.12$\times$ \\
128K & 1 & 2.90$\times$ & 3.58$\times$ & 4.47$\times$ & 3.55$\times$ & 3.21$\times$ & 2.84$\times$ \\
32K & 4 & 1.35$\times$ & 0.68$\times$ & 1.36$\times$ & 0.96$\times$ & 0.99$\times$ & 0.99$\times$ \\
64K & 4 & 2.12$\times$ & 1.24$\times$ & 2.12$\times$ & 2.00$\times$ & 1.79$\times$ & 2.19$\times$ \\
128K & 4 & 2.83$\times$ & 2.05$\times$ & 3.85$\times$ & 2.39$\times$ & 3.57$\times$ & 3.18$\times$ \\
\bottomrule
\end{tabular}
\end{table}

\FloatBarrier
\section{Configuration Sensitivity}
\label{app:sensitivity}

\subsection{Block size}

Table~\ref{tab:block_sensitivity} compares $B\in\{64,128,256\}$ after recalibrating each setting to approximately 5.34\% actual density. At the reported operating points, $B=128$ gives the highest RBS accuracy. Smaller blocks increase the number of block summaries and candidate pairs to process, while larger blocks coarsen the grouping of relevant and irrelevant tokens. The observed accuracy depends on the selector as well as the granularity: RBS is below centroid-only at $B=64$ and $B=256$ in this sweep. Thus, the evidence supports $B=128$ as the selected configuration rather than a claim of robustness to arbitrary block size. These sensitivity scores are separate from the 130-example selector comparison in Table~\ref{tab:selector_ablation}.

\begin{table}[htbp]
\centering
\caption{Block-size sensitivity on RULER-128K with Qwen3-30B-A3B-Instruct-2507-FP8. Each setting is recalibrated to the 5.34\% density anchor. Density and accuracy are percentages. The reference RBS $B=128$ row is the primary budget-sweep operating point.}
\label{tab:block_sensitivity}
\begin{tabular}{@{}lrrrr@{}}
\toprule
Selector & Block size & Calibration parameter & Actual density & Accuracy \\
\midrule
Centroid-only & 64 & 0.100 & 5.302 & 74.17 \\
& 128 & 0.194 & 5.300 & 74.73 \\
& 256 & 0.400 & 5.276 & 72.15 \\
\midrule
RBS union & 64 & 0.096 & 5.343 & 71.86 \\
& 128 & Reference & 5.340 & 80.28 \\
& 256 & 0.382 & 5.361 & 70.73 \\
\bottomrule
\end{tabular}
\end{table}

\subsection{Thresholds and model transfer}

Table~\ref{tab:threshold_sensitivity} applies the same three pairs of thresholds to the MoE and dense Qwen3 models. The \emph{conservative} and \emph{aggressive} labels refer to observed retention, not guaranteed accuracy: the conservative setting retains more blocks, and the aggressive setting retains fewer. Fixed thresholds yield different actual densities across models. In particular, the default pair produces 4.997\% mean density on Qwen3-30B-A3B and 6.029\% on Qwen3-32B in this diagnostic. This is expected for content-dependent selection and motivates reporting measured density with every configuration.

Accuracy need not vary monotonically with density when both branch thresholds change: they alter which blocks are selected as well as how many. The cross-model diagnostic demonstrates this distinction and should be interpreted within its own evaluation rather than compared directly with the full primary benchmark scores.

\begin{table}[htbp]
\centering
\caption{Supplementary cross-model threshold sensitivity on RULER-128K. Density and accuracy are percentages. The same threshold pairs are used for both models; the radius normalization adapts to each prompt at runtime.}
\label{tab:threshold_sensitivity}
\small
\setlength{\tabcolsep}{5pt}
\begin{tabular}{@{}llrrrrr@{}}
\toprule
Model & Setting & $\alpha_{\rm base}$ & $\alpha_{\rm rescue}$ & Mean density & p90 density & Accuracy \\
\midrule
Qwen3-30B-A3B & Conservative & 0.16 & 0.24 & 6.478 & 7.282 & 75.31 \\
FP8 & Default & 0.22 & 0.18 & 4.997 & 5.441 & 71.86 \\
& Aggressive & 0.24 & 0.16 & 4.676 & 5.048 & 75.12 \\
\midrule
Qwen3-32B & Conservative & 0.16 & 0.24 & 7.983 & 8.845 & 65.05 \\
& Default & 0.22 & 0.18 & 6.029 & 6.561 & 63.90 \\
& Aggressive & 0.24 & 0.16 & 5.600 & 6.078 & 61.74 \\
\bottomrule
\end{tabular}
\end{table}

\FloatBarrier
\section{Calibration, Runtime Adaptation, and Memory}
\label{app:deployment}

\subsection{Offline calibration and online selection}

Table~\ref{tab:calibration} quantifies calibration cost on 50 prompts of length 128K. RBS uses three configurations, taking 0.443 wall-clock hours and 0.885 GPU-hours with TP=2. The reported maximum error from its 5\% target is 0.0078 percentage point. The median/90th-percentile choices for the radius normalization remain fixed throughout calibration.

Calibration is an offline operating-point choice. At inference time, the thresholds are fixed and the prompt/layer/head-specific radius distribution directly determines $\beta_b$. Table~\ref{tab:dynamic_cost} reports total selector time at the 128K system operating point, including statistics, normalization, and selection. The RBS selector takes 694.81 ms, including its two-branch scoring and mask construction; this work is part of the measured execution path. Relative to centroid-only selection, the extra work implements the adaptive rescue policy evaluated in the quality experiments.

\begin{table}[htbp]
\centering
\caption{Offline calibration on 50 prompts of length 128K using the A100 main-model setup (TP=2). GPU-hours are the reported aggregate device time; small rounding differences from twice the wall time are retained. Maximum density error is in percentage points (pp). RBS fixes the radius quantiles at 0.5/0.9 while calibrating its selection thresholds.}
\label{tab:calibration}
\small
\begin{tabular}{@{}lrrrrr@{}}
\toprule
Method & Configurations & Wall hours & GPU-hours & Target density & Max. error (pp) \\
\midrule
RBS-Attention & 3 & 0.443 & 0.885 & 5\% & 0.0078 \\
FlashPrefill & 3 & 0.418 & 0.837 & 5\% & 0.1625 \\
MInference & 3 & 0.232 & 0.464 & 2/5/10\% & 0.0913 \\
\bottomrule
\end{tabular}
\end{table}

\begin{table}[htbp]
\centering
\caption{Total per-request selector time at the 128K A100 system operating point. Times include block statistics, normalization, score computation, threshold/Top-K decisions, and index construction; RBS additionally unions its branch masks. There is no per-request threshold search.}
\label{tab:dynamic_cost}
\begin{tabular}{@{}lr@{}}
\toprule
Selector & Total selector time (ms) \\
\midrule
FlashPrefill & 349.89 \\
Full-L2 & 279.21 \\
RBS-Attention & 694.81 \\
Quest-style & 1,195.88 \\
\bottomrule
\end{tabular}
\end{table}

\subsection{Persistent statistics, workspace, and model peak}

Table~\ref{tab:metadata} reports storage for block statistics and the temporary selector workspace. With the reported accounting dimensions of 1,024 blocks, head dimension 128, and four KV heads, centroid-only statistics occupy 1.000 MiB per layer. Adding a FP32 radius and a FP32 adaptive coefficient increases RBS's statistics to 1.031 MiB per layer, whereas the two BF16 vectors used by Quest-style selection require 2.000 MiB. This is a comparison of block-statistic representations, not of total allocator usage.

The temporary workspace is substantially larger than these statistics. RBS uses 1,155.601 MiB in the reported implementation, compared with 515.499 MiB for centroid-only and 428.231 MiB for Quest-style selection. Thus, the small persistent radius overhead does not imply an equally small total selector footprint. Table~\ref{tab:peak_memory} measures the separate full-model quantity: at 128K, the RBS run peaks at 60.486 GiB allocated, compared with 61.552 GiB for Dense. These measurements include the complete serving configuration and depend on allocator and workspace lifetimes; they are not a claim that sparse prefill compresses or evicts the KV cache.

\begin{table}[htbp]
\centering
\caption{Selector storage under the reported accounting dimensions: 1,024 blocks, head dimension 128, four KV heads, and 48 active layers. Statistics are reported per layer; temporary workspace is reported separately and should not be multiplied by the layer count. Vectors use BF16 and the radius/$\beta$ scalars use FP32.}
\label{tab:metadata}
\begin{tabular}{@{}llrr@{}}
\toprule
Selector & Statistics per block & MiB/layer & Workspace (MiB) \\
\midrule
Centroid-only & 128-vector & 1.000 & 515.499 \\
Full-L2 & 128-vector + radius & 1.016 & 515.324 \\
RBS-Attention & 128-vector + radius + $\beta$ & 1.031 & 1,155.601 \\
Quest-style & Two 128-vectors & 2.000 & 428.231 \\
\bottomrule
\end{tabular}
\end{table}

\begin{table}[htbp]
\centering
\caption{Full-model peak memory on A100 GPUs with Qwen3-30B-A3B-Instruct-2507-FP8 (TP=2). Each cell reports \textbf{max-rank allocated / reserved} memory in GiB. The largest tested length in this supplementary memory experiment is 128K; it is not a measured hardware capacity limit.}
\label{tab:peak_memory}
\begin{tabular}{@{}lccc@{}}
\toprule
Method & 32K & 64K & 128K \\
\midrule
Dense & 55.710 / 56.650 & 57.328 / 59.279 & 61.552 / 65.510 \\
Centroid-only & 55.248 / 57.010 & 56.865 / 58.918 & 61.090 / 63.441 \\
RBS-Attention & 54.644 / 56.357 & 56.262 / 58.266 & 60.486 / 64.016 \\
Quest-style & 59.894 / 61.541 & 61.512 / 63.449 & 61.330 / 63.496 \\
\bottomrule
\end{tabular}
\end{table}

\FloatBarrier
\section{Geometric Bounds and the Dual-Branch Selector}
\label{app:geometry}

This appendix separates the geometric motivation for the radius from the selection policy that uses it. Throughout, $K_b$ is a finite set of key vectors, $c_b$ is its mean, $r_b=\max_{k\in K_b}\|k-c_b\|_2$, and $q$ is a query. We omit the common attention scale $1/\sqrt{d}$ because it does not affect the inequalities.

\subsection{The full L2 bound and the attenuated score}

For every $k\in K_b$,
\begin{align}
    q^\top k
    &=q^\top c_b+q^\top(k-c_b)\\
    &\leq q^\top c_b+\|q\|_2\|k-c_b\|_2\\
    &\leq q^\top c_b+\|q\|_2r_b.
\end{align}
Taking the maximum over $k$ proves Eq.~\eqref{eq:full_l2_bound}. Geometrically, this is the maximum linear score over a ball containing all keys in the block.

The actual rescue score replaces $r_b$ by $\beta_b r_b$. When $\beta_b<1$, the smaller ball need not contain the keys. For example, take $K_b=\{-u,u\}$ with $\|u\|_2=1$ and $q=u$. Then $c_b=0$, $r_b=1$, and the largest token logit is $1$, whereas the rescue logit is $\beta_b$. This provides a direct counterexample to interpreting the attenuated score as a universal upper bound. The radius supplies a geometric risk signal; quantile-based attenuation controls how strongly that signal influences selection.

\subsection{Comparison with a coordinate-wise box}

For coordinate $j$, let $a_{b,j}=\min_{k\in K_b} k_j$ and $u_{b,j}=\max_{k\in K_b} k_j$. The page score used by Quest~\citep{tang2024quest} is
\begin{equation}
    U_b^{\mathrm{box}}(q)
    =\sum_{j=1}^{d}\max(q_j a_{b,j},q_j u_{b,j}).
    \label{eq:quest_box}
\end{equation}
Each coordinate of every key lies in its corresponding interval, so $\max_{k\in K_b}q^\top k\leq U_b^{\mathrm{box}}(q)$. Equivalently, if $h_b=(u_b-a_b)/2$ and $v_b=(u_b+a_b)/2$, then $U_b^{\mathrm{box}}(q)=q^\top v_b+|q|^\top h_b$. The box center $v_b$ need not equal the key mean $c_b$.

The two bounds have no uniform ordering. Two-dimensional examples make this explicit:
\begin{itemize}
    \item Let $K=\{e_1,-e_1,e_2,-e_2\}$ and $q=(1,1)^\top$. The centroid is zero and the radius is one, giving $U^{\mathrm{L2}}(q)=\sqrt{2}$ and $U^{\mathrm{box}}(q)=2$. Here the L2 bound is tighter: the box contains the corner $(1,1)$ even though no key lies there.
    \item Let $K=\{e_1,-e_1,\epsilon e_2,-\epsilon e_2\}$ for $0<\epsilon<1$, and let $q=e_2$. The radius is again one, giving $U^{\mathrm{L2}}(q)=1$ and $U^{\mathrm{box}}(q)=\epsilon$. Here the box bound is tighter because it captures the narrow range in the second coordinate.
\end{itemize}
These examples illustrate the trade-off between a coordinate-wise envelope and an isotropic one. Under a common orthogonal transformation of keys and queries, the centroid dot product and L2 radius are unchanged. A coordinate-wise box can change under the same transformation. Rotation invariance is a geometric property, not a guarantee that the L2 selector will have better accuracy or a tighter bound on a given input.

\subsection{Why separate thresholds matter}

An upper bound describes the possible score within one block. Selecting a fixed number of blocks by upper bound does not by itself guarantee recovery of the globally highest-scoring keys: a loose bound can rank an irrelevant block above a relevant one. Likewise, the radius-adaptive score does not imply a guarantee of preserving dense-attention mass. We evaluate these selection decisions empirically.

\method{} makes two distinct decisions for each query block: which candidates are sufficiently relevant by centroid score, and which candidates merit retention after radius-based uplift. With a single rescue score, the uplift of one candidate can raise the relative threshold and remove other candidates, even though none of their individual logits decreased. The independently computed base mask prevents that loss for candidates retained by the centroid rule. The final union can therefore differ from both centroid-only selection and a single full-L2 selector.

The thresholds are controls on relative score, not exact cardinality constraints. If the same candidates and scores are held fixed, decreasing either threshold cannot shrink its branch's mask. Across prompts and layers, however, the score distribution and branch overlap vary, so fixed thresholds need not yield the same density. This is why comparisons at a common nominal setting and comparisons after calibrating actual density answer different experimental questions.

\end{document}